\documentclass[10pt,twocolumn,letterpaper]{article}

\usepackage{cvpr}
\usepackage{times}
\usepackage{epsfig}
\usepackage{graphicx}
\usepackage{amsmath}
\usepackage{amssymb}
\usepackage{caption}

\usepackage[breaklinks=true,bookmarks=false]{hyperref}

\cvprfinalcopy 

\def\cvprPaperID{6621} 

\ifcvprfinal\fi

\begin{document}

\renewcommand{\thefootnote}{\fnsymbol{footnote}}

\title{Distilling Knowledge from Graph Convolutional Networks}

\author{
Yiding Yang$^1$,
Jiayan Qiu$^2$,
Mingli Song$^3$,
Dacheng Tao$^2$,
Xinchao Wang$^1$\footnotemark[2] \\ 
$^1$Department of Computer Science, Stevens Institute of Technology, USA\\
$^2$School of Computer Science, Faculty of Engineering, The University of Sydney, Australia\\
$^3$College of Computer Science and Technology, Zhejiang University, China\\
{\tt\small
\{yyang99, xwang135\}@stevens.edu,
jqiu3225@uni.sydney.edu.au,}\\
{\tt\small
brooksong@zju.edu.cn,
dacheng.tao@sydney.edu.au
}
}

\twocolumn[{%
\renewcommand\twocolumn[1][]{#1}%
\maketitle
\thispagestyle{empty}
\vspace{-1.0cm}
\begin{center}
    \centering
    \includegraphics[width=0.9\textwidth]{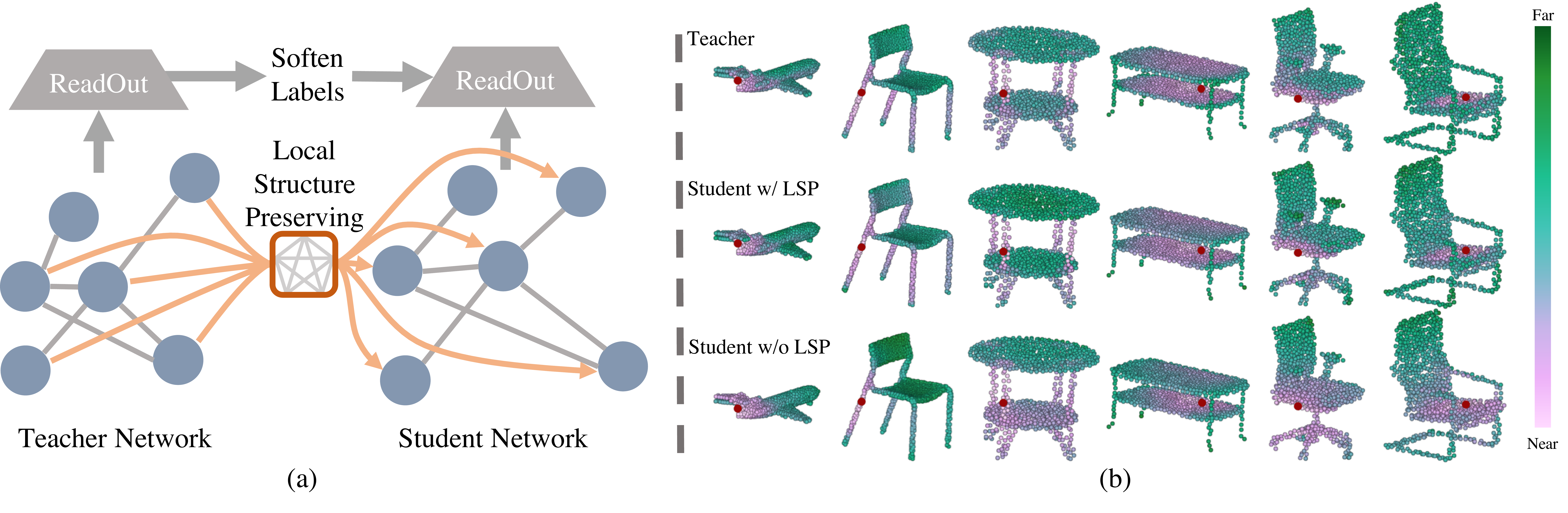}
    \vspace{-2mm}
    \captionof{figure}{(a) Unlike existing 
    knowledge distillation methods that  focus on only the prediction or the middle activation, our method explicitly distills knowledge about how the teacher model embeds
    the topological structure and transfers it to the student model. 
    (b) We display the structure of the feature space, visualized
    by the distance between the red point and the others on a \textbf{point cloud} dataset.
    Here, each object is represented as a set of 3D points.
   \textbf{Top Row}: structures obtained from the teacher; \textbf{Middle Row}: structures  obtained from the student trained with the local structure preserving~(LSP) module; \textbf{Bottom Row}: structures  obtained from the student trained without LSP.  
    Features in the middle and bottom row    are obtained from the last layer of the model after training 
    for ten epochs. 
   As we can see, model trained with LSP learns a similar structure as that of the teacher, while the model without LSP fails to do so.
    }
    \label{fig:ituition}
\end{center}
}]

\footnotetext[2]{Corresponding author.}

\begin{abstract}
   Existing knowledge distillation methods focus on 
   convolutional neural networks~(CNNs), where the input samples
   like images  lie in a grid domain, and have largely
    overlooked graph convolutional networks~(GCN)
   that handle non-grid data. 
    In this paper, we propose to our best knowledge 
    the first dedicated approach to {distilling} knowledge
   from a pre-trained GCN model. 
   To enable the knowledge transfer from the teacher GCN to
   the student, we propose a local structure preserving module
   that explicitly accounts for the topological semantics 
   of the teacher. In this module,   
    the local structure information from both the
   teacher  and the student  are extracted
   as distributions, and hence minimizing the distance between these
   distributions enables topology-aware knowledge transfer from the teacher, 
   yielding a compact yet high-performance student model. 
   Moreover, the proposed approach is readily extendable 
   to dynamic graph models, 
   where the input graphs for the teacher and the student
   may differ.
  We evaluate the proposed method on two different datasets
   using GCN models of different architectures, and demonstrate that
   our method achieves the state-of-the-art knowledge
   distillation performance for GCN models.
   Code is publicly available at \url{https://github.com/ihollywhy/DistillGCN.PyTorch}.
   
\end{abstract}
\vspace{-4mm}

\section{Introduction}

Deep neural networks~(DNNs) have demonstrated
their unprecedented results in almost all computer vision tasks.
The state-of-the-art performances, however, come at the cost
of the very high computation and memory loads,
which in many cases preclude the deployment of DNNs
on the edge side. 
To this end, knowledge distillation has been proposed, which is
one of the main streams of model 
compression~\cite{hinton_distill,dodeepnets,accelerate_cnn,LowrankCompress}.
By treating a pre-trained cumbersome network as the
teacher model, knowledge distillations aims 
to learn a compact student
model, which is expected  to 
master the expertise of the teacher,
via transferring knowledge from the teacher.






The effectiveness of knowledge distillation has been validated in many tasks,
where the performance of the student closely approaches that of the teacher.
Despite the encouraging progress, existing knowledge distillation schemes 
have been focusing on 
convolutional neural networks~(CNNs),
for which the input samples, such as images, 
lie in the grid domain.
However, many real-life data, such as point clouds,
take the form of non-grid structures like graphs
and thus call for the 
graph convolutional networks~(GCNs)~\cite{gat,nmessagepassing,gcn,graphsage}.
GCNs explicitly looks into
the topological structure of the
data by exploring the local and global semantics of the graph.
As a result, conventional knowledge distillation methods,
which merely account for the output
or the intermediate activation  and 
omit the topological context of input data,
are no longer capable to fully carry out the knowledge transfer.

In this paper, we introduce to our best knowledge 
the first dedicated knowledge distillation approach tailored
for GCNs. Given a pre-trained teacher GCN,
our goal is to train a student GCN model with fewer layers,
or  lower-dimension feature maps, or even  a smaller graph
with fewer edges. 
At the heart of our GCN distillation is the capability to encode the topological
information concealed in the graph,
which is absent in prior CNN-based methods. 
As depicted in Fig.~\ref{fig:ituition}, 
the proposed method considers the features of node 
as well as the topological connections among them,
and hence provides the student model richer 
and more critical information about topological structure embedded 
by the teacher.



\begin{figure*}[t]
\begin{center}
    \includegraphics[width=0.82\linewidth]{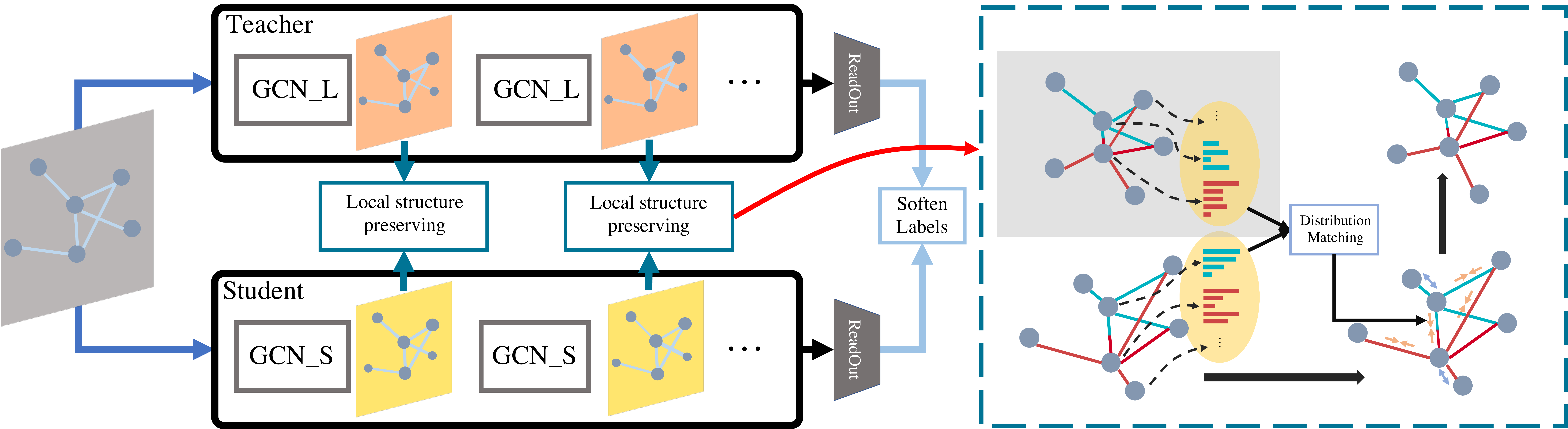}
\end{center}
\vspace{-3mm}
   \caption{Framework of the proposed knowledge distillation method for GCNs. The local structure preserving module is
   the core of the proposed method. Given the feature maps and the graphs from the teacher and the student, we first compute the distribution of the local structure for each node and then match the distributions of the teacher 
   with that of the student. The student model will be optimized by minimizing the difference of distribution among all the local structures.}
\label{fig:framework}
\vspace{-4mm}
\end{figure*}


We illustrate the workflow of the 
proposed GCN knowledge distillation approach in Fig.~\ref{fig:framework}.
We design a local structure preserving~(LSP) module
to explicitly account for the graphical semantics. 
Given the embedded feature of node and 
the graph from teacher and student, 
the LSP module measures the topological difference 
between them and guides the student to obtain a similar topological embedding
as the teacher does. Specifically, LSP
first generates distribution for each local structure from both the student and the teacher,
and then enforces the student to learn a similar local structure by minimizing the distance
between the distributions. 
Furthermore, our approach can be readily 
extended to  dynamic graph models, where the graphs are not
static but constructed by the teacher and student model dynamically.

To see the distillation performance, 
we evaluate our method on two 
different tasks in different domains include Protein-Protein Interaction dataset for node classification and ModelNet40 for 3D object recognition.
The models used in these two tasks are in different architectures.
Experiments show that our method consistently achieves 
the best knowledge distillation 
performance  among all the compared methods,
validating the effectiveness and generalization of 
the proposed approach.

Our contributions are summarized as following:
\begin{itemize}
    \item We introduce a novel method for distilling knowledge from graph convolutional network.
    To the best of our knowledge, this is the first dedicated knowledge distillation method tailored for GCN models.
    
    \item We devise a local structure preserving~(LSP) method to measure the similarity of the
    local topological structures embedded by teacher and student, enabling our method to be 
    readily extendable to dynamic graph models.
    
    \item We evaluate the proposed method on two different tasks of different domains and on GCN models of different architectures, and show that our method outperforms all other methods consistently.
\end{itemize}

\section{Related Work}

Many methods have been proposed to distill the knowledge 
from a trained model and transfer it to a student model with smaller capacity~\cite{dodeepnets,accelerate_cnn,yim2017gift}.
Although there are many distillation strategies
that do not only utilize the output~\cite{hinton_distill}
but also focus on the intermediate activation~\cite{fitnets,paying_attention,like_what_you_like},
they are all designed for the deep convolutional network
with the grid data as input.

Our method, however, focuses on the graph convolutional networks, which   
handles a more general input that lies in the non-grid domain. 
To the best of our knowledge, this is the first
attempt along this line.
In what follows, we briefly review several tasks that are related to our method.

\textbf{Knowledge Distillation. }
Knowledge Distillation~(KD) is first proposed in~\cite{hinton_distill}, 
where the goal is distilling the knowledge from a teacher model 
that is typical large into a smaller model so that
the student model can hold a similar performance as the teacher's.
In this method, the output of teacher is smoothed by setting
a high temperature in the softmax function, which make it contains
the information of the relationship among classes.
Beside the output, intermediate activation can also be utilized
to better train a student network. FitNet~\cite{fitnets} forces the 
student to learn a similar features as teacher's by adding an additional
fully connected layer to transfer features of student model.
~\cite{paying_attention} proposes a method to transfer the attention instead
of the feature itself to get a better distillation performance.
Moreover, NST~\cite{like_what_you_like} provides a method to 
learn a similar activation of the neurons. There are also many other methods
in this field~\cite{liu2019knowledgeIRG,dodeepnets,accelerate_cnn,binarizeddistill,liu2019knowledge,chen2019datafree,mean_teacher,yim2017gift,chen2015net2net,chen2019datafree}, but none of them provide a solution that is suitable for GCN.

\textbf{Knowledge Amalgamation. }
Knowledge amalgamation~\cite{shen2019customizing,knowledge_flow,shen2019amalgamating,ye2019amalgamating} 
aims to learn a student network from multiple teachers from different domains. 
The student model is trained as a multi-task model and learns from all the teachers.
For example, a method of knowledge amalgamation is proposed in~\cite{ye2019student_master} 
to train a student model from heterogeneous-task teachers include scene parsing teacher,
depth estimation teacher and surface-normal estimation teacher. The student model
will have a backbone network trained from all the teachers' knowledge and also
several heads for different tasks trained from the corresponding teacher's knowledge. 
MTZ~\cite{he2018multi} is a framework to compress multiple 
but correlated models into one model. The knowledge is distilled
via a layer-wise neuron sharing mechanism.
CFL~\cite{luo2019knowledge} distills the knowledge by learning a common feature space, 
wherein the student model mimics the transformed features of the teachers to aggregate knowledge. 
Although many such methods are proposed, the models involved
are usually limited within grid domain.

\textbf{Graph Convolutional Network. }
In recent years, graph convolutional network~\cite{splinecnn,tgcn,gatedgcn,Monet,nmessagepassing,ijcai2019-spagan}
has been proved to be a powerful
model for non-grid data, which is typically represented as a set of nodes with
features along with a graph that represents the relationship among nodes.
The first GCN paper~\cite{gcn} shows that the GCN can be built by
making the first-order approximation of spectral graph convolutions.
A huge amount of methods have been proposed to
make the GCN more powerful.
GraphSAGE~\cite{graphsage} gives a solution to make the GCN model
scalable for huge graph by sampling the neighbors rather than using 
all of them. GAT~\cite{gat} introduces the attention mechanism to GCN
to make it possible to learn the weight for each neighbors automatically.
\cite{2018adaptive} improves the efficient of training by adaptive sampling.
In this paper, instead of designing a new GCN, we focus on how to
transfer the knowledge effectively between different GCN models.

\textbf{3D Object Recognition. }
One of the setups for 3D object recognition is predicting the label
of the object given a set of 3D points belong to it~\cite{qi2017pointnet,qi2017pointnet++}.
Deep learning based methods~\cite{wu20153d_deeppointcloud,guo20153d_deeppointcloud,fang20153d_deeppointcloud,li2016fpnn_deeppointcloud,qi2016volumetric_deeppointcloud,maturana2015voxnet_deeppointcloud}
outperform the previous methods that based on hand-crafted feature extractors~\cite{aubry2011wave_handcraftedPointcloud,bronstein2010scale_handcraftedPointcloud}.
Moreover, GCN based
methods~\cite{feng2019hypergraph,li2018pointcnn,wang2018dynamic,landrieu2018large,yang2018foldingnet},
which can directly encode the structure information from the set of points, become one of the most
popular directions along this line.
Graph in these methods are typically obtained by
connecting the k nearest points,
where the distance is measured in the original
space~\cite{landrieu2018large} or the learned feature space~\cite{wang2018dynamic}.

\section{Method}

In this section, we first give a brief description about the 
GCN followed by the motivation
of the proposed knowledge distillation method that is
based on the observation of the fundamental mechanism of GCN.
We then provide the details about the local structure preserving~(LSP)
module, which is the core of our proposed method.
Moreover, we explore the different choices of the distance functions
used in the LSP module.
Finally, we give the scheme to extend LSP
to the dynamic graph models.

\subsection{Graph Convolutional Network}
Unlike the traditional convolutional networks that take grid data as 
input and output the high-level features, the input of graph convolutional
networks can be non-grid, which is more general. Such non-grid 
input data is
typically represented as a set of features 
$\mathbb{X} = \{x_1,x_2,...,x_n\}\in \mathbb{R}^F$,
and a directed/undirected graph $\mathcal{G}=\{\mathcal{V}, \mathcal{E}\}$.
For example, in the task of 3D object recognition, we can set $x_i$ as
the 3D coordination and $\mathcal{E}$ as the set of the nearest neighbors.

Given the input $X$ and $\mathcal{G}$, the core operation of graph convolutional network is shown
as:
\begin{equation}
    x'_i =  \mathcal{A}_{j:(j,i)\in \mathcal{E}} h_\mathbf{\theta} 
    (g_\mathbf{\phi}(x_i), g_\mathbf{\phi}(x_j)),
    \label{eq:aggregate}
\end{equation}
where $h_\mathbf{\theta}$ is a function 
that considers the features in pair wise, 
$g_\mathbf{\phi}$  is a function to map the features into
a new space,
$\mathcal{A}$ is the strategy of how to aggregate features from the neighbors
and get the new feature of the center node $i$.

There are many choices of function $h$, function $g$ and the aggregation strategies.
Take the graph attention network~\cite{gat} as an example.
it can be formulated as
\begin{equation}
    x'_i = \sum_{j:{j,i}\in \mathcal{E}}
     \frac{ e^{MLP_1(x_i || x_j)} }
        { \sum_{j:(j,i)\in \mathcal{E}}(e^{MLP_1(x_i || x_j)})  }
    MLP_2(x_i),
    \label{eq:gat}
\end{equation}
where the function $g$ is designed as a multilayer perceptron, 
function $h$ is another multilayer perceptron that takes
pair of nodes as input and predict the attention between them.
The aggregation strategy is weighted summing all the features of neighbors according 
to the attention after normalization.

\subsection{Motivation}
The motivation of the proposed method is based on the
fundamental of graph convolutional network. As shown in Eq.~\ref{eq:aggregate},
the aggregation strategy~($\mathcal{A}$) plays an important role in
embedding the features of nodes~\cite{gatedgcn,gat}, 
which is learned during the training process.
We thus aim to provide the student the information
about the function that the teacher has learned.
However, it is challenging to distill knowledge that exactly
represents the aggregation function and transfer it to the student.
Instead of distilling the aggregation function directly, 
we distill the outcomes of such function: the embedded topological
structure. The student  can then be guided by matching 
the structure embedded by itself and that
embedded by the teacher.
We will show in the following sections how to describe
the topological structure information and distill it to 
the student.

\subsection{Local Structure Preserving}
For the intermediate feature maps of a GCN, we can formulate
it as a graph $\mathcal{G}=\{\mathcal{V},\mathcal{E}\}$ 
and a set of features $\mathbb{Z}=\{z_1,z_2,...,z_n\}\in \mathbb{R}^F$,
where $n$ is the number of nodes, $F$ is the dimension of the feature maps.
The local structure can be summarized
as a set of vectors $\mathbb{LS}=\{LS_1, LS_2,...,LS_n\}, LS_i\in \mathbb{R}^d$, 
where $d$ is the degree of the center node $i$
of the local structure.
Each element of the vector is computed by
\begin{equation}
    LS_{ij} = \frac{e^{\mathcal{SIM}(z_i, z_j)}}{\sum_{j:(j,i)\in \mathcal{E}}
    ( e^{\mathcal{SIM}(z_i, z_j)} )},
    \label{eq:lsp}
\end{equation}
\begin{equation}
    \mathcal{SIM}(z_i, z_j) = || z_i - z_j||_2^2.
    \label{eq:sim_1}
\end{equation}
where $\mathcal{SIM}$ is a function that measures the similarity of
the given pair of nodes, which can be defined as the 
euclidean distance between the two features. There are
also many other advanced functions can be used here, which we will
give more details in the following section.
We take an exponential operation
and normalize the values 
across all the nodes that point to center of the local structure.
As a result, for each node $i$, we can obtain its corresponding 
local structure representation $LS_i \in \mathbb{R}^d$ by applying Eq.~\ref{eq:lsp}. 
Notice that for different center node, their
local structure representation may be in different dimension, 
which is depended on it's local graph.

In the setting of knowledge distillation, we are given a teacher network
as well as a student one, where the teacher network is trained and fixed.
We first provide here a local structure preserving strategy under
the situation that both these two networks take the same graph as input
but with different layers and dimension of embedded features.
For the dynamic graph models, where the graph can be changed during the
optimization process, we will give a solution in the section~\ref{dynamicgraph}.

Given the intermediate feature maps, we can compute the
local structure vectors for both the teacher and the student
networks, which are donated as $\mathcal{LS}^s$ and $\mathcal{LS}^t$. 
For each center node $i$, the similarity of the local structure
between the student's and the teacher's can be computed as
\begin{equation}
    \mathcal{S}_i = D_{KL}(LS_i^s||LS_i^t) 
                = \sum_{j:(j,i)\in \mathcal{E}}
                LS_{ij}^s log(\frac{LS_{ij}^s}{LS_{ij}^t}),
\end{equation}
where the Kullback Leibler divergence is adopted.

A smaller $\mathcal{S}_i$ means a more similar distribution of the local structure.
Thus, we compute the similarity of the distributions over all the nodes of 
the given graph and obtain the local structure preserving loss as
\begin{equation}
    \mathcal{L}_{LSP} = \frac{1}{N} \sum_{i=1}^N \mathcal{S}_i.
\end{equation}

The total loss is formulated as:
\begin{equation}
    \mathcal{L} = \mathcal{H}(p_s, y) + \lambda \mathcal{L}_{LSP}
\end{equation}
where $y$ is the label and $p_s$ is the prediction of the student model,
$\lambda$ is the hyperparameter to balance these two losses and $\mathcal{H}$
represents cross entropy loss function that is also adopted by many other
knowledge distillation methods~\cite{fitnets,like_what_you_like,paying_attention}.

\subsection{Kernel Function}
The similarity measurement function shown in Eq.~\ref{eq:sim_1} makes a strong assumption 
about the feature space that the similarity between pair of nodes is proportional 
to their euclidean distance, which is typically not the truth. 
We thus apply the kernel trick, which is very useful to map the vectors
to a higher dimension and compute the similarity, to address this problem.

Kernel tricks are widely used in the traditional statistical machine learning methods. 
In that situation, the original feature vector will be mapped to a 
higher dimension by a implicit function $\varphi$. The similarity of the feature vector
will then be computed as the inner product of the two mapped vectors 
as $\left\langle \varphi(z_i), \varphi(z_j) \right\rangle$. By adopting the kernel function, we can compute the above
two steps together without knowing the expression of $\varphi$.

There are several choices of the kernel functions. Three of the most common used kernel functions
are linear function~(Linear), polynomial kernel function~(Poly) and radial basis function~(RBF) kernel:
\begin{equation}
    K(z_i, z_j) = 
    \begin{cases}
        (z_i^T z_j+c)^d &                               \text{Poly}\\
        e^{-\frac{1}{2\sigma^2} ||z_i-z_j||^2 } &     \text{RBF} \\
        z_i^T z_j &                               \text{Linear}\\
    \end{cases}
    \label{eq:kernelfunctions}
\end{equation}
In this paper, we adopt and compare these three kernel functions
and also the $L_2$ Norm. For the polynomial kernel function, $d$ and $c$ are
set to two and zero respectively. For the RBF, $\sigma$ is set to one.

\begin{figure}[t]
\begin{center}
    \includegraphics[width=0.9\linewidth]{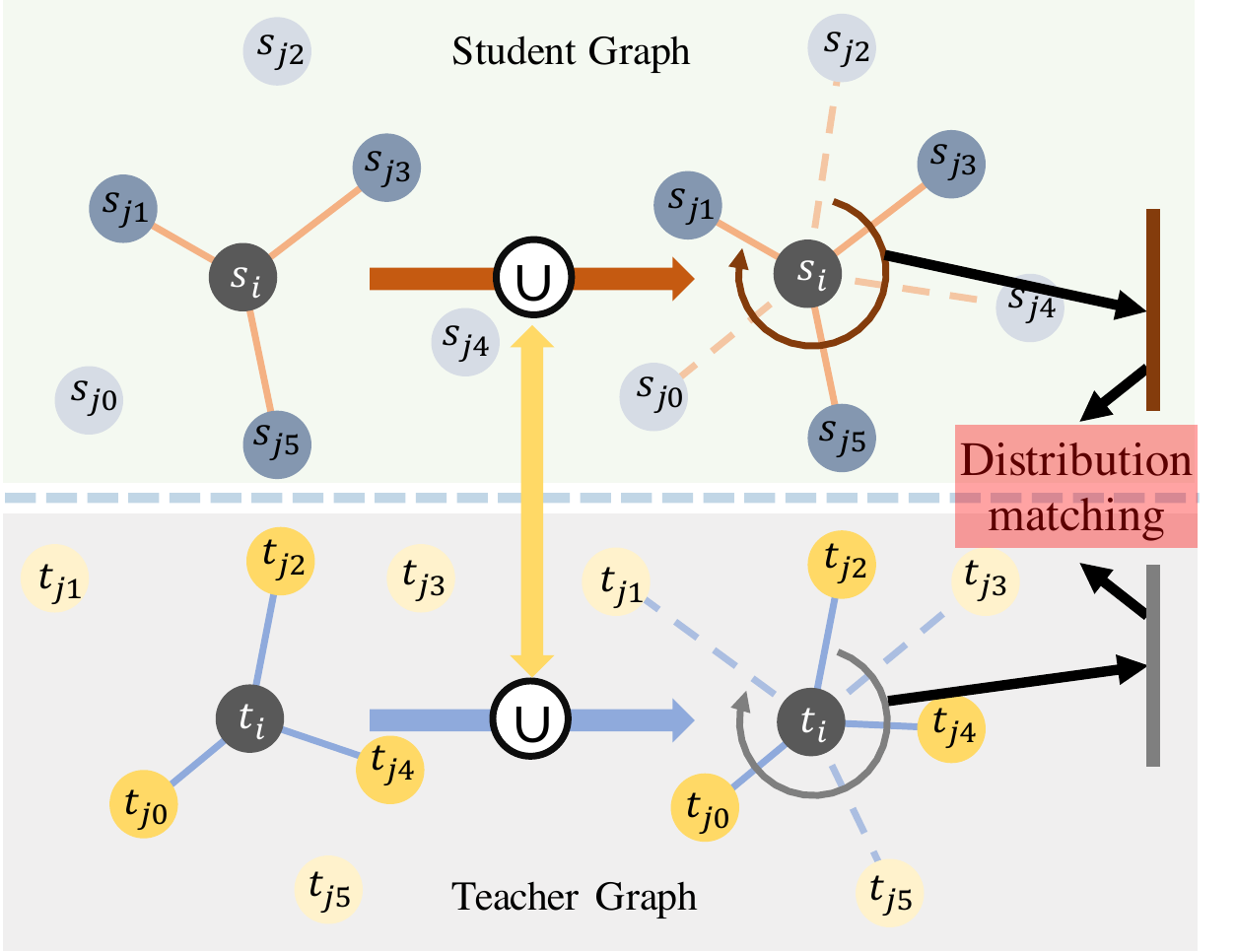}
\end{center}
\vspace{-5mm}
   \caption{Handling models with a dynamic graph, where the graph may be updated during the training process. We address this problem by first adding virtual edges according to the union of the two graphs
   from the student model and teacher model.
   The local structure preserving module can be then applied to the new graph directly.}
\label{fig:dynamicgraph}
\vspace{-3.5mm}
\end{figure}

\subsection{Dynamic Graph} \label{dynamicgraph}
While the above method can be applied to the GCN models that with a fixed graph as 
input, it lacks flexibility for dynamic graph models.
In the setup of dynamic graph model, both the feature of the nodes and the connection
among them can be changed. Take the DGCNN model~\cite{wang2018dynamic} as an example,
the graph is initially constructed according to the 3D coordination of the input nodes/points
and will be reconstructed once a new feature of nodes are obtained.

The advantage of dynamic graph is that the graph can represent the topological
connection in a learned feature space rather than always in the initial feature space.
However, such dynamic graph method can cause a problem to the local structure preserving module
we mentioned above. The left of Fig.~\ref{fig:dynamicgraph} is a case of the intermediate
graphs generated by the DGCNN model. For each layer, the graph 
is constructed by finding the $K$ closet nodes for each node, where $K$ is the hyperparameter.
Directly computing the local structure vector using above method is meaningless 
because the distributions will come from a total different order of nodes.

We proposed a strategy to deal with such kind of situation by adding virtual edges
to the graph in both the teacher model and student one.
As shown in Fig.~\ref{fig:dynamicgraph}, given the two graphs constructed by the 
teacher and the student, which typically do not hold the same distribution of edges,
we obtain the union of edges 
$\mathcal{E}^{u}_{\bullet i}=\{(j,i): (j,i) \in \mathcal{E}^t | (j,i)\in \mathcal{E}^s \}$ 
for each center node $i$.
A similar local structure vector can be obtained by replacing $\mathcal{E}$ with 
$\mathcal{E}^u$ in Eq.~\ref{eq:lsp}. Notice that although we adding the virtual edges
for both the graphs, we only use them for local structure preserving module.
The models still use their original graph to aggregate and update the feature of nodes.
By considering the union structure of these two graphs, the generated local structure
vectors now involve the nodes with the same distribution, which makes it possible for
the student network to learn the topological relationship learned by the teacher.

Such strategy not only make it possible to compare the embedded local structure with
different distribution of neighbors but also compare between local structure with 
different size of neighbors. It means that we can distill the knowledge from a teacher
model with a large $K$ to the student model with smaller $K$.

\section{Experiments}

In this section, we first give a brief description about the comparison methods.
Then, we provide the experimental setup include the datasets we used, the GCN models
we adopted and the details for each comparison methods.
Notice that our goal is not to achieve the state-of-the-art performance in each
dataset or task but to transfer as mush as information from the teacher model
the student one. This can be measured by the performance of the student model when
the same teacher pre-trained model is involved.

We adopt two datasets in different domains, One is the protein-protein interaction~(PPI)
~\cite{ppi}
dataset where the graphs come from the human tissues. This is a common used
dataset for node classification~\cite{gat,graphsage}. Another dataset is
ModelNet40~\cite{modelnet40} that
contains point clouds come from the CAD models, which is a common used
dataset for 3D object 
analysis~\cite{qi2017pointnet,qi2017pointnet++,li2018pointcnn,wang2018dynamic}.

We also evaluate the proposed knowledge distillation method on GCN models
with different architectures. Specifically, for the PPI dataset, GAT~\cite{gat} 
model is adopted that takes fixed graph as input. 
For the ModelNet40 dataset, DGCNN~\cite{wang2018dynamic} model with dynamic
graph is adopted.
We show in the experiments that our proposed method achieve the state-of-the-art
distillation performance under various setups.

\subsection{Comparison Methods}
Since there is no knowledge distillation method designed for 
GCN models, we implement three knowledge distillation methods that
can be used directly in a GCN model includes KD method~\cite{hinton_distill},
FitNet method~\cite{fitnets} and
attention transfer method~(AT)~\cite{paying_attention}.
Neuron selectivity transfer~\cite{like_what_you_like}
is also one of the knowledge distillation methods for traditional convolutional
networks. We leave it out since it makes an assumption that the size of
the feature maps of student's and teacher's should be in the same size,
which is not the case of our setup for the student and teacher models.
Besides all the methods, we also set a baseline method which is training the student
model with the original loss.

The summary of the comparison methods is as follow:
\begin{itemize}
    \item \textbf{KD method~\cite{hinton_distill}} is the first attempt for distilling knowledge
    from a teacher network. It utilizes a soften labels generated by the teacher network
    as an additional supervision. The intuition behind this method is that the soften label 
    contains the similarity information among the classes learned by the teacher
    network. Since this method only relies on the output, it is suitable for most kinds
    of models.
    \item \textbf{FitNet method~\cite{fitnets}} does not only make use of the output of the teacher model
    but also consider the intermediate feature maps. This method based on an assumption
    that the feature of the teacher model can be recovered from the feature of the
    student when the student is well trained. It introduces an additional mapping function
    to map the feature of student's to that of the teacher's and compute the $L_2$ distance
    between the mapped feature and the teacher's feature.
    \item \textbf{Attention transfer method~(AT)~\cite{paying_attention}} 
    provides another way to transfer the knowledge in attention
    domain. In this method, the student model is forced to focus on the similar spatial areas
    like the teacher does, which is achieved by adding an $L_2$ loss between their
    attention maps. The attention map can be obtained from the feature maps 
    and keep in the same size without consideration of the different channels.
\end{itemize}

\begin{figure*}[t]
\begin{center}
    \includegraphics[width=0.82\linewidth]{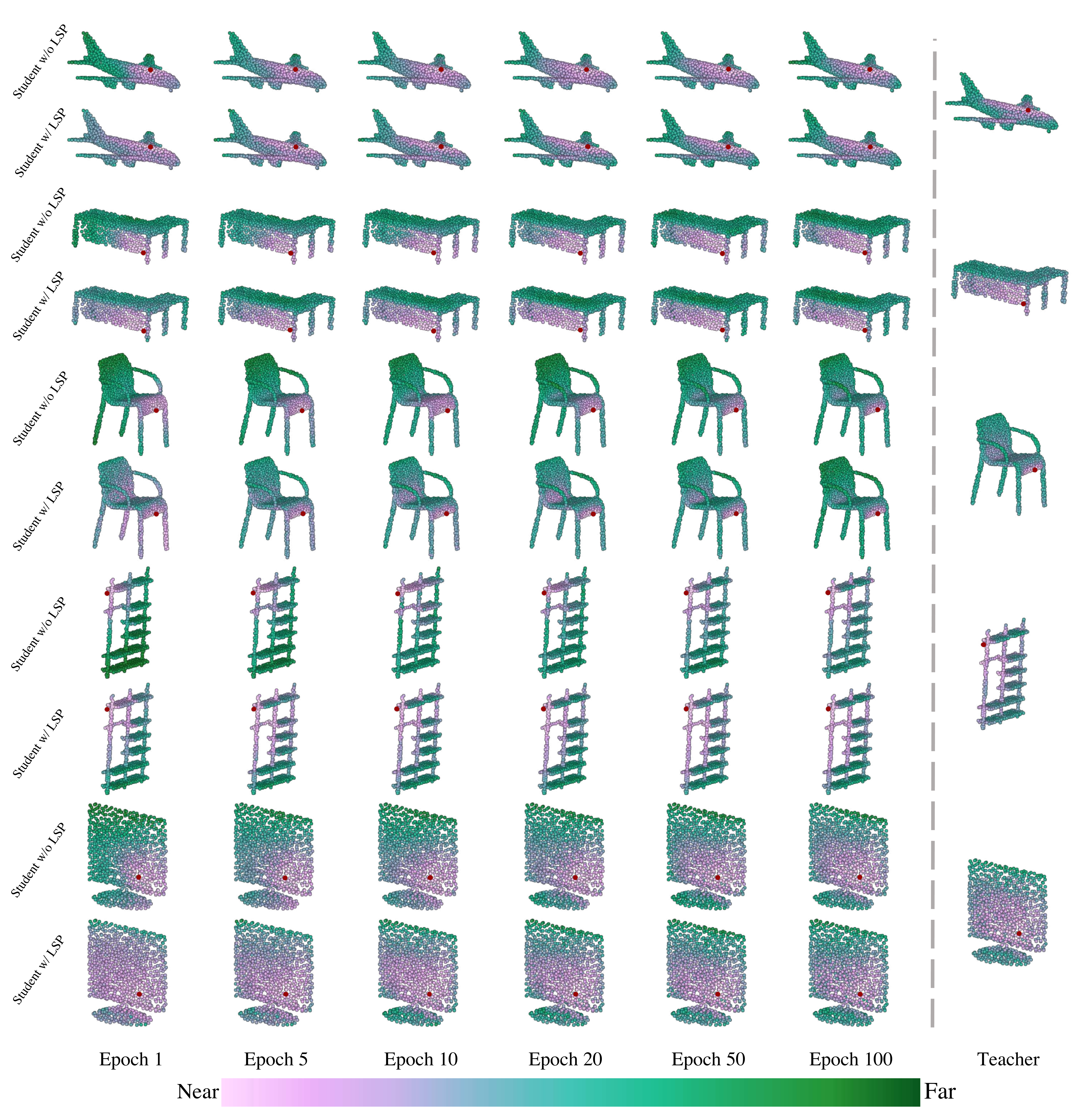}
\end{center}
\vspace{-3mm}
   \caption{Visualization of the structure of the learned feature space, depicted by 
   the distance between the red point and others. The 
   features are extracted from the last layer of models. From left to right, we show 
   the structure from the model trained in epoch 1, 5, 10,
   20, 50, and 100. The \textbf{rightmost column} are obtained from the teacher model. For each object, 
   the \textbf{lower row} is obtained from student model trained with our proposed method,
   and the \textbf{upper row} is obtained from student model trained with cross entropy loss. 
   Our proposed knowledge distillation method  guides the student model
   to embed the local structure as the teacher model does, leading to a similar 
   structure in the very early training stage.}
   \vspace{-3mm}
\label{fig:bigvis}
\end{figure*}

\subsection{Node Classification}
In the node classification task, we are given the input nodes with associated features
and also the graph. The goal is to generate the embedded feature for each node
such that nodes with different classes can be separated. We adopt protein-protein interaction~(PPI)
dataset that contains 24 graphs corresponding to different human tissues. We follow the same
dataset splitting protocol, wherein 20 graphs are used for training, two graphs are used for validation and
another two graphs are used for testing. 
The average number of nodes for each graph in this dataset is 2372 
and each node has an average degree of 14. 
The dimension of input feature of the nodes is 50 and the number of class is 121.

For this dataset and task, we adopt GAT model for both the teacher model and
student model. Since this is a multi-label task, where each node
can belong to more than one classes, the binary cross entropy loss is adopted.
The architecture of these two models are shown as follow:
\begin{table}[h!]
\centering
\scalebox{0.9}{
 \begin{tabular}{c | c c  c} 
 \hline
 \textbf{Model} & \textbf{Layers} & \textbf{Attention heads} & \textbf{Hidden features} \\
 \hline 
 Teacher & 3 & 4,4,6 & 256,256,121\\ 
 Student & 5 & 2,2,2,2,2 & 68,68,68,68,121 \\
 \hline
 \end{tabular}
 }
 \vspace{-2mm}
 \caption{Summary of the teacher and student models used on the 
 PPI dataset for node classification. 
 The student network is deeper than the teacher 
 but with lower dimension of hidden features.}
 \vspace{-5mm}
\end{table}

\begin{table}[h!]
\centering
\scalebox{0.8}{
 \begin{tabular}{c | c c c | c} 
 \hline
 \textbf{Model} & \textbf{Params} & \textbf{RunTime} & \textbf{Training} & \textbf{F1 Score} \\
 \hline 
 Teacher                                & 3.64M & 48.5ms & 1.7s/3.4G & 97.6 \\ 
 \hline
 Student\_Full                          & 0.16M & 41.3ms & 1.3s/1.2G & 95.7 \\
 Student\_KD~\cite{hinton_distill}      & - & - & - & - \\
 Student\_AT~\cite{paying_attention}    & 0.16M & 41.3ms & 1.9s/1.4G & 95.4 \\
 Student\_FitNet~\cite{fitnets}         & 0.16M & 41.3ms & 2.4s/1.6G & 95.6 \\
 Student\_LSP~(Ours)                    & 0.16M & 41.3ms & 2.0s/1.5G & \textbf{96.1} \\ [1ex] 
 \hline
 \end{tabular}
 }
 \vspace{-2mm}
 \caption{Node classification results on the PPI dataset. 
 Teacher model used on this dataset is a GAT model with three hidden layers.
 full means the student model is trained with the ground truth labels
without the teacher model.}
 \label{tab:nodeclassification}
 \vspace{-3mm}
\end{table}

The results are shown in Tab.~\ref{tab:nodeclassification}.
Params represents the total number of parameters;
RunTime is the inference time for one sample and Training is the training time/GPU memory usage for one iteration, which is measured in
a Nvidia 1080Ti GPU.
The optimizer, learning rate, weight decay and training epochs
are set to Adam, 0.005, 0 and 500 respectively for all the methods.
All other hyperparameters for each method are tuned to obtain the best results on the validation set.
Specifically, for AT method, the attention is computed by as
$\sum_{i=1}^{C}{|F_i|}$; 
For ours, the kernel function is set to RBF and $\lambda$ is set to 100.
Notice that the loss function does not involve the softmax function.
Therefore, the KD method, which is used by setting a high temperature in a softmax function, 
is not suitable here.

As what can be seen, all the knowledge distillation methods expect ours
fail to provide a positive influence to the student model and lead to a 
drop of the performance respect to the performance of model 
trained with the original loss. Our method, thanks to the ability to transfer
the local structure information to the student model, provides a positive
influence and lead to a student model with the best performance
among all the comparison methods.

\subsection{3D Object Recognition}
We adopt ModelNet40~\cite{modelnet40} dataset for the 3D object recognition task, 
wherein the object is represented by a set of points with only the 3D coordination as features. 
There are 40 classes on this dataset and objects in each class
come from the CAD models.

In this experiment, the architecture of both the teacher and student models is the same as DGCNN~\cite{wang2018dynamic}.
DCGNN is a dynamic graph convolutional model 
that makes both the advantages of the PointNet~\cite{qi2017pointnet} and 
graph convolutional network. The graph here is constructed according to the distance of the
points in their feature space. Since the feature space is different in different layers and
different training stages, the graph is also changed, making it a dynamic graph convolutional
model. 

The teacher model is under the same setup as the original paper: it has five graph convolutional
layers followed by two fully connected layers. The student model has four graph convolutional
layers with fewer feature map channels followed by one fully connected layers.
The size of graph used in the student model,
which is determined by the number of neighbors~(K) when constructing the graph, 
is also smaller than the teacher's.
The details of these two models are summarized in Tab.~\ref{3dmodes}.

\begin{table}[h!]
\centering
\scalebox{0.87}{
 \begin{tabular}{c | c c c  c} 
 \hline
 \textbf{Model} & \textbf{Layers} & \textbf{Feature map's size} & \textbf{MLPs} &  \textbf{K}\\
 \hline
 Teacher & 5 & 64,64,128,256,1024 & 512,256 & 20\\ 
 Student & 4 & 32,32,64,128 & 256 & 10 \\
 \hline
 \end{tabular}
 }
 \vspace{-2mm}
 \caption{Summary of the teacher and student models used on the
 ModelNet40 dataset. The student network is with less layers,
 fewer channels, and smaller input graphs.}
\vspace{-2mm}
 \label{3dmodes}
\end{table}

The results are shown in Tab.~\ref{3Dresults}.
For the KD method, $\alpha$
is set to 0.1, the same as the original paper.
The optimizer, learning rate, momentum and training epoch are set to
SGD, 0.1, 0.9, 250 respectively for all the comparison methods. 
The other hyperparameters for each method are turned to obtain the 
best accuracy on the validation set.
For our method, the kernel function is set to RBF and $\lambda$ is set to 100.

We can see from the results that both KD method and AT method can boost
the performance of the student model. Our method, thanks to the ability
to learn the structure information from dynamic graph,
generates the best student model for both the accuracy and mean class accuracy.
The student model with only fix percent parameters can achieve a similar performance
as the teacher model, which shows the generalization ability of the proposed method 
for different tasks and different architectures.

\begin{table}[h!]
\centering
\scalebox{0.75}{
 \begin{tabular}{c | c c c | c c} 
 \hline
 \textbf{Model} & \textbf{Params} & \textbf{RunTime} & \textbf{Training} & \textbf{Acc} & \textbf{mAcc} \\
 \hline
 Teacher                                & 1.81M & 8.72ms & 30s/4.2G & 92.4 & 89.3\\ 
 \hline
 Student\_Full                          & 0.1M & 3.31ms &  12s/1.4G & 91.2 & 87.5\\
 Student\_KD~\cite{hinton_distill}      & 0.1M & 3.31ms &  15s/1.4G & 91.6 & 88.1\\
 Student\_AT~\cite{paying_attention}    & 0.1M & 3.31ms &  21s/1.7G & 91.6 & 87.9\\
 Student\_FitNet~\cite{fitnets}         & 0.1M & 3.31ms &  28s/2.4G & 91.1 & 87.9 \\
 Student\_LSP~(Ours)                    & 0.1M & 3.31ms &  29s/2.2G & \textbf{91.9} & \textbf{88.6}\\ [1ex] 
 \hline
 \end{tabular}
 }
 \vspace{-2mm}
 \caption{3D object recognition results on ModelNet40. 
 Teacher network used here is a DGCNN model with four graph
 convolutional layers.}
 \label{3Dresults}
\vspace{-3mm}
\end{table}

\subsection{Structure Visualization}
In order to provide an intuitive understanding of the proposed method,  
we visualize the structure of the learned feature space during the process of
optimization, which is represented by
the distance among points in the object.
As shown in Fig.~\ref{fig:bigvis}, the student model trained with the 
proposed method~(shown in the lower row for each object)
can learn a similar structure as the teacher model very quickly
in the early training stage. It can partially explain why the proposed
method can generate a better student model.

\subsection{Ablation and Performance Studies}
To thoroughly evaluate our method, we provide here 
ablation and performance studies
include the influence of different kernel functions as well as
the performance for different student model configurations.
All the experiments for this section are conducted on the ModelNet40
dataset.

\textbf{Different Kernel Functions.}
We test all the three different kernel functions as Eq.~\ref{eq:kernelfunctions} 
and also the naive $L_2$ Norm. 
The results are shown in Tab.~\ref{tab:kernel}.
All the functions can provide the positive information to get a 
better student model and RBF works the best.

\begin{table}[h!]
\centering
\scalebox{0.87}{
 \begin{tabular}{c | c c } 
 \hline
 \textbf{Model} & \textbf{Acc} & \textbf{mAcc}  \\
 \hline
 LSP w/ $L_2$ Norm & 91.4 & 88.3 \\ 
 LSP w/ Polynomial function & 91.7 & 87.7 \\
 LSP w/ RBF        & \textbf{91.9} & \textbf{88.6} \\
 LSP w/ Linear function     & 91.5 & 87.8 \\
 \hline
 \end{tabular}
 }
 \vspace{-2mm}
 \caption{Performance with different kernel functions. RBF achieves 
 the overall best performance.}
 \label{tab:kernel}
\vspace{-3mm}
\end{table}

\textbf{Different Model Configurations.}
We provide here the experiments to evaluate the trade-off between the
model complexity and the performance when training with our proposed
method.
Specifically, we change the student model by adding more channels
to each layer, adding one more graph convolutional layers
and adding one more fully connected layers.
Notice that student model with more channels holds almost the 
same accuracy performance as the teacher model but with less many parameters.
\begin{table}[h!]
\centering
\scalebox{0.8}{
 \begin{tabular}{c | c c | c c } 
 \hline
 \textbf{Model} & \textbf{Params~(M)} & \textbf{RunTime~(ms)} & \textbf{Acc} & \textbf{mAcc}  \\
 \hline
 W/ more Channels   & 0.44 & 3.83 & 92.3 & 88.7 \\ 
 W/ more Layers     & 0.14 & 4.26 & 92.1 & 89.2 \\
 W/ more MLPs       & 0.30 & 3.67 & 91.8 & 88.6 \\
 \hline
 \end{tabular}
 }
 \vspace{-2mm}
 \caption{Performance with different model configurations to 
 evaluate the trade-off between performance and model-size/run-time.}
\vspace{-3mm}
\end{table}

\section{Conclusion}
In this paper,  we propose a dedicated approach to
distilling knowledge from GCNs, which is to our best knowledge the first
attempt along this line.
This is achieved by preserving the local structure of the teacher network
during the training process.
We represent the local structure of the intermediate feature
maps as distributions over the similarities between the center node
of the local structure and its neighbors, so that preserving
the local structures are equivalent to matching the distributions.
Moreover, the proposed approach can be readily extended to 
dynamic graph models.
Experiments on two datasets in different domains
and on two GCN models of different architectures
demonstrate that the proposed method yields state-of-the-art distillation performance,
outperforming  existing knowledge distillation methods. 

\section*{Acknowledgement}
This work is supported by Australian
Research Council Projects FL-170100117, DP-180103424 and Xinchao Wang's startup funding of Stevens Institute of Technology.

\newpage

{\small
\bibliographystyle{ieee_fullname}
\bibliography{gcndistill}

\begin{thebibliography}{10}\itemsep=-1pt

\bibitem{aubry2011wave_handcraftedPointcloud}
Mathieu Aubry, Ulrich Schlickewei, and Daniel Cremers.
\newblock The wave kernel signature: A quantum mechanical approach to shape
  analysis.
\newblock In {\em 2011 IEEE international conference on computer vision
  workshops (ICCV workshops)}, pages 1626--1633. IEEE, 2011.

\bibitem{dodeepnets}
Jimmy Ba and Rich Caruana.
\newblock Do deep nets really need to be deep?
\newblock In {\em Advances in neural information processing systems}, pages
  2654--2662, 2014.

\bibitem{bronstein2010scale_handcraftedPointcloud}
Michael~M Bronstein and Iasonas Kokkinos.
\newblock Scale-invariant heat kernel signatures for non-rigid shape
  recognition.
\newblock In {\em 2010 IEEE Computer Society Conference on Computer Vision and
  Pattern Recognition}, pages 1704--1711. IEEE, 2010.

\bibitem{chen2019datafree}
Hanting Chen, Yunhe Wang, Chang Xu, Zhaohui Yang, Chuanjian Liu, Boxin Shi,
  Chunjing Xu, Chao Xu, and Qi Tian.
\newblock Data-free learning of student networks.
\newblock {\em arXiv preprint arXiv:1904.01186}, 2019.

\bibitem{chen2015net2net}
Tianqi Chen, Ian Goodfellow, and Jonathon Shlens.
\newblock Net2net: Accelerating learning via knowledge transfer.
\newblock {\em arXiv preprint arXiv:1511.05641}, 2015.

\bibitem{tgcn}
Jian Du, Shanghang Zhang, Guanhang Wu, Jos{\'e}~MF Moura, and Soummya Kar.
\newblock Topology adaptive graph convolutional networks.
\newblock {\em arXiv preprint arXiv:1710.10370}, 2017.

\bibitem{fang20153d_deeppointcloud}
Yi Fang, Jin Xie, Guoxian Dai, Meng Wang, Fan Zhu, Tiantian Xu, and Edward
  Wong.
\newblock 3d deep shape descriptor.
\newblock In {\em Proceedings of the IEEE Conference on Computer Vision and
  Pattern Recognition}, pages 2319--2328, 2015.

\bibitem{feng2019hypergraph}
Yifan Feng, Haoxuan You, Zizhao Zhang, Rongrong Ji, and Yue Gao.
\newblock Hypergraph neural networks.
\newblock In {\em Proceedings of the AAAI Conference on Artificial
  Intelligence}, volume~33, pages 3558--3565, 2019.

\bibitem{splinecnn}
Matthias Fey, Jan Eric~Lenssen, Frank Weichert, and Heinrich M{\"u}ller.
\newblock Splinecnn: Fast geometric deep learning with continuous b-spline
  kernels.
\newblock In {\em Proceedings of the IEEE Conference on Computer Vision and
  Pattern Recognition}, pages 869--877, 2018.

\bibitem{nmessagepassing}
Justin Gilmer, Samuel~S Schoenholz, Patrick~F Riley, Oriol Vinyals, and
  George~E Dahl.
\newblock Neural message passing for quantum chemistry.
\newblock In {\em Proceedings of the 34th International Conference on Machine
  Learning-Volume 70}, pages 1263--1272. JMLR. org, 2017.

\bibitem{guo20153d_deeppointcloud}
Kan Guo, Dongqing Zou, and Xiaowu Chen.
\newblock 3d mesh labeling via deep convolutional neural networks.
\newblock {\em ACM Transactions on Graphics (TOG)}, 35(1):3, 2015.

\bibitem{graphsage}
Will Hamilton, Zhitao Ying, and Jure Leskovec.
\newblock Inductive representation learning on large graphs.
\newblock In {\em Advances in Neural Information Processing Systems}, pages
  1024--1034, 2017.

\bibitem{he2018multi}
Xiaoxi He, Zimu Zhou, and Lothar Thiele.
\newblock Multi-task zipping via layer-wise neuron sharing.
\newblock In {\em Advances in Neural Information Processing Systems}, pages
  6016--6026, 2018.

\bibitem{hinton_distill}
Geoffrey Hinton, Oriol Vinyals, and Jeffrey Dean.
\newblock Distilling the knowledge in a neural network.
\newblock In {\em NIPS Deep Learning and Representation Learning Workshop},
  2015.

\bibitem{2018adaptive}
Wenbing Huang, Tong Zhang, Yu Rong, and Junzhou Huang.
\newblock Adaptive sampling towards fast graph representation learning.
\newblock In {\em Advances in Neural Information Processing Systems}, pages
  4558--4567, 2018.

\bibitem{like_what_you_like}
Zehao Huang and Naiyan Wang.
\newblock Like what you like: Knowledge distill via neuron selectivity
  transfer.
\newblock {\em arXiv preprint arXiv:1707.01219}, 2017.

\bibitem{gcn}
Thomas~N Kipf and Max Welling.
\newblock Semi-supervised classification with graph convolutional networks.
\newblock {\em arXiv preprint arXiv:1609.02907}, 2016.

\bibitem{landrieu2018large}
Loic Landrieu and Martin Simonovsky.
\newblock Large-scale point cloud semantic segmentation with superpoint graphs.
\newblock In {\em Proceedings of the IEEE Conference on Computer Vision and
  Pattern Recognition}, pages 4558--4567, 2018.

\bibitem{li2018pointcnn}
Yangyan Li, Rui Bu, Mingchao Sun, Wei Wu, Xinhan Di, and Baoquan Chen.
\newblock Pointcnn: Convolution on x-transformed points.
\newblock In {\em Advances in Neural Information Processing Systems}, pages
  820--830, 2018.

\bibitem{li2016fpnn_deeppointcloud}
Yangyan Li, S{\"o}ren Pirk, Hao Su, Charles~R Qi, and Leonidas~J Guibas.
\newblock Fpnn: Field probing neural networks for 3d data.
\newblock In {\em Advances in Neural Information Processing Systems}, pages
  307--315, 2016.

\bibitem{gatedgcn}
Yujia Li, Daniel Tarlow, Marc Brockschmidt, and Richard Zemel.
\newblock Gated graph sequence neural networks.
\newblock {\em arXiv preprint arXiv:1511.05493}, 2015.

\bibitem{knowledge_flow}
Iou-Jen Liu, Jian Peng, and Alexander Schwing.
\newblock Knowledge flow: Improve upon your teachers.
\newblock In {\em International Conference on Learning Representations}, 2019.

\bibitem{liu2019knowledgeIRG}
Yufan Liu, Jiajiong Cao, Bing Li, Chunfeng Yuan, Weiming Hu, Yangxi Li, and
  Yunqiang Duan.
\newblock Knowledge distillation via instance relationship graph.
\newblock In {\em Proceedings of the IEEE Conference on Computer Vision and
  Pattern Recognition}, pages 7096--7104, 2019.

\bibitem{liu2019knowledge}
Yufan Liu, Jiajiong Cao, Bing Li, Chunfeng Yuan, Weiming Hu, Yangxi Li, and
  Yunqiang Duan.
\newblock Knowledge distillation via instance relationship graph.
\newblock In {\em Proceedings of the IEEE Conference on Computer Vision and
  Pattern Recognition}, pages 7096--7104, 2019.

\bibitem{luo2019knowledge}
Sihui Luo, Xinchao Wang, Gongfan Fang, Yao Hu, Dapeng Tao, and Mingli Song.
\newblock Knowledge amalgamation from heterogeneous networks by common feature
  learning.
\newblock {\em arXiv preprint arXiv:1906.10546}, 2019.

\bibitem{maturana2015voxnet_deeppointcloud}
Daniel Maturana and Sebastian Scherer.
\newblock Voxnet: A 3d convolutional neural network for real-time object
  recognition.
\newblock In {\em 2015 IEEE/RSJ International Conference on Intelligent Robots
  and Systems (IROS)}, pages 922--928. IEEE, 2015.

\bibitem{Monet}
Federico Monti, Davide Boscaini, Jonathan Masci, Emanuele Rodola, Jan Svoboda,
  and Michael~M Bronstein.
\newblock Geometric deep learning on graphs and manifolds using mixture model
  cnns.
\newblock In {\em Proceedings of the IEEE Conference on Computer Vision and
  Pattern Recognition}, pages 5115--5124, 2017.

\bibitem{qi2017pointnet}
Charles~R Qi, Hao Su, Kaichun Mo, and Leonidas~J Guibas.
\newblock Pointnet: Deep learning on point sets for 3d classification and
  segmentation.
\newblock In {\em Proceedings of the IEEE Conference on Computer Vision and
  Pattern Recognition}, pages 652--660, 2017.

\bibitem{qi2016volumetric_deeppointcloud}
Charles~R Qi, Hao Su, Matthias Nie{\ss}ner, Angela Dai, Mengyuan Yan, and
  Leonidas~J Guibas.
\newblock Volumetric and multi-view cnns for object classification on 3d data.
\newblock In {\em Proceedings of the IEEE conference on computer vision and
  pattern recognition}, pages 5648--5656, 2016.

\bibitem{qi2017pointnet++}
Charles~Ruizhongtai Qi, Li Yi, Hao Su, and Leonidas~J Guibas.
\newblock Pointnet++: Deep hierarchical feature learning on point sets in a
  metric space.
\newblock In {\em Advances in neural information processing systems}, pages
  5099--5108, 2017.

\bibitem{fitnets}
Adriana Romero, Nicolas Ballas, Samira~Ebrahimi Kahou, Antoine Chassang, Carlo
  Gatta, and Yoshua Bengio.
\newblock Fitnets: Hints for thin deep nets.
\newblock {\em arXiv preprint arXiv:1412.6550}, 2014.

\bibitem{shen2019amalgamating}
Chengchao Shen, Xinchao Wang, Jie Song, Li Sun, and Mingli Song.
\newblock Amalgamating knowledge towards comprehensive classification.
\newblock In {\em Proceedings of the AAAI Conference on Artificial
  Intelligence}, volume~33, pages 3068--3075, 2019.

\bibitem{shen2019customizing}
Chengchao Shen, Mengqi Xue, Xinchao Wang, Jie Song, Li Sun, and Mingli Song.
\newblock Customizing student networks from heterogeneous teachers via adaptive
  knowledge amalgamation.
\newblock {\em arXiv preprint arXiv:1908.07121}, 2019.

\bibitem{mean_teacher}
Antti Tarvainen and Harri Valpola.
\newblock Mean teachers are better role models: Weight-averaged consistency
  targets improve semi-supervised deep learning results.
\newblock In {\em Advances in neural information processing systems}, pages
  1195--1204, 2017.

\bibitem{gat}
Petar Veli{\v{c}}kovi{\'c}, Guillem Cucurull, Arantxa Casanova, Adriana Romero,
  Pietro Lio, and Yoshua Bengio.
\newblock Graph attention networks.
\newblock {\em arXiv preprint arXiv:1710.10903}, 2017.

\bibitem{binarizeddistill}
Haoyu Wang, Defu Lian, and Yong Ge.
\newblock Binarized collaborative filtering with distilling graph convolutional
  networks.
\newblock {\em arXiv preprint arXiv:1906.01829}, 2019.

\bibitem{wang2018dynamic}
Yue Wang, Yongbin Sun, Ziwei Liu, Sanjay~E Sarma, Michael~M Bronstein, and
  Justin~M Solomon.
\newblock Dynamic graph cnn for learning on point clouds.
\newblock {\em arXiv preprint arXiv:1801.07829}, 2018.

\bibitem{accelerate_cnn}
Zhenyang Wang, Zhidong Deng, and Shiyao Wang.
\newblock Accelerating convolutional neural networks with dominant
  convolutional kernel and knowledge pre-regression.
\newblock In {\em European Conference on Computer Vision}, pages 533--548.
  Springer, 2016.

\bibitem{wu20153d_deeppointcloud}
Zhirong Wu, Shuran Song, Aditya Khosla, Fisher Yu, Linguang Zhang, Xiaoou Tang,
  and Jianxiong Xiao.
\newblock 3d shapenets: A deep representation for volumetric shapes.
\newblock In {\em Proceedings of the IEEE conference on computer vision and
  pattern recognition}, pages 1912--1920, 2015.

\bibitem{modelnet40}
Zhirong Wu, Shuran Song, Aditya Khosla, Fisher Yu, Linguang Zhang, Xiaoou Tang,
  and Jianxiong Xiao.
\newblock 3d shapenets: A deep representation for volumetric shapes.
\newblock In {\em Proceedings of the IEEE conference on computer vision and
  pattern recognition}, pages 1912--1920, 2015.

\bibitem{yang2018foldingnet}
Yaoqing Yang, Chen Feng, Yiru Shen, and Dong Tian.
\newblock Foldingnet: Point cloud auto-encoder via deep grid deformation.
\newblock In {\em Proceedings of the IEEE Conference on Computer Vision and
  Pattern Recognition}, pages 206--215, 2018.

\bibitem{ijcai2019-spagan}
Yiding Yang, Xinchao Wang, Mingli Song, Junsong Yuan, and Dacheng Tao.
\newblock Spagan: Shortest path graph attention network.
\newblock In {\em Proceedings of the Twenty-Eighth International Joint
  Conference on Artificial Intelligence, {IJCAI-19}}, pages 4099--4105.
  International Joint Conferences on Artificial Intelligence Organization, 7
  2019.

\bibitem{ye2019student_master}
Jingwen Ye, Yixin Ji, Xinchao Wang, Kairi Ou, Dapeng Tao, and Mingli Song.
\newblock Student becoming the master: Knowledge amalgamation for joint scene
  parsing, depth estimation, and more.
\newblock In {\em Proceedings of the IEEE Conference on Computer Vision and
  Pattern Recognition}, pages 2829--2838, 2019.

\bibitem{ye2019amalgamating}
Jingwen Ye, Xinchao Wang, Yixin Ji, Kairi Ou, and Mingli Song.
\newblock Amalgamating filtered knowledge: Learning task-customized student
  from multi-task teachers.
\newblock {\em arXiv preprint arXiv:1905.11569}, 2019.

\bibitem{yim2017gift}
Junho Yim, Donggyu Joo, Jihoon Bae, and Junmo Kim.
\newblock A gift from knowledge distillation: Fast optimization, network
  minimization and transfer learning.
\newblock In {\em Proceedings of the IEEE Conference on Computer Vision and
  Pattern Recognition}, pages 4133--4141, 2017.

\bibitem{LowrankCompress}
Xiyu Yu, Tongliang Liu, Xinchao Wang, and Dacheng Tao.
\newblock On compressing deep models by low rank and sparse decomposition.
\newblock In {\em The IEEE Conference on Computer Vision and Pattern
  Recognition (CVPR)}, July 2017.

\bibitem{paying_attention}
Sergey Zagoruyko and Nikos Komodakis.
\newblock Paying more attention to attention: Improving the performance of
  convolutional neural networks via attention transfer.
\newblock {\em arXiv preprint arXiv:1612.03928}, 2016.

\bibitem{ppi}
Marinka Zitnik and Jure Leskovec.
\newblock Predicting multicellular function through multi-layer tissue
  networks.
\newblock {\em Bioinformatics}, 33(14):i190--i198, 2017.

\end{thebibliography}
}

\end{document}